# Pareto-optimal lane-changing motion planning in mixed traffic

Yang Li, Linbo Li, Daiheng Ni

*Abstract*— **This paper applies the pareto-optimal concept to LC (lane-changing) motion planning in the presence of mixed traffic including manual and autonomous vehicles. Firstly, a multiobjective optimization problem is presented, in which the comfort, efficiency and safety of the LC vehicle and the surrounding vehicles are jointly modelled. Thereafter, the pareto-optimal solutions are obtained through employing the NSGA-II (Non-dominated Sorting Genetic -II) algorithm. Finally, the experiment section analyzes the (macroscopic and microscopic) lane-changing impact from a pareto-optimal perspective. Also, a comprehensive sensitivity analysis is conducted. Our results demonstrate that our algorithm could significantly reduce the lane-changing impact within its region, and the total costs are reduced in the range of 10.94% to 48.66%. This paper could be considered as a preliminary research framework for the application of the pareto-optimal concept. We hope this research will provide valuable insights into autonomous driving technology.**

*Index Terms*—Lane-changing motion planning, Pareto-optimal front, Mixed traffic, Longitudinal control model.

## I. INTRODUCTION

A LONG with Car-following (CF) behavior, lane-changing (LC) behavior is also an indispensable component of traffic flow theories [1], which describes the lateral movement of the vehicle from current lane to target lane while proceeding forward. LC behavior can be divided into mandatory and discretionary. Lane changes due to diversion, merging or avoiding collision with obstacles can be viewed as mandatory. Lane changes resulting from driver's need for higher speed or more comfortable driving space can be considered as discretionary. For the convenience of discussion, we present a typical scenario in Fig. 1. In the most complex scenario, lane change will involve interaction with up to four vehicles.

Research in lane change can be divided into decision-making [2, 3], motion planning [2, 4-9], trajectory prediction [10], duration analysis [11-13], and impact analysis [14-16]. The research on decision-making mainly addresses the question of whether perform lane-changing or lane keeping. The research could be categorized as famous Gipps-type [17], utility theory [3] type, game theory [18], deep learning [4], etc. The research in LC impact mainly estimate, model and alleviate the effect of LC on its surroundings [15, 16, 19, 20]. The studies in duration analysis mainly answers the question about the distribution and affecting factors of lane-changing duration [11]. This paper is concerned with LC motion planning, which is an indispensable component research of lane change.

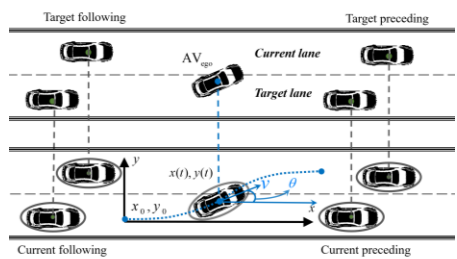

**Fig. 1 A typical LC schematic where five vehicles are simultaneously involved**

Trajectory planning and tracking are the two main components of motion planning research. When a vehicle has determined to perform lane change, the vehicle would gradually turn the steering wheel and drive towards the target lane. More specifically, trajectory planning part calculates a designed trajectory in advance, and trajectory tracking controls a vehicle to drive along a trajectory until it arrives the center-line of the target lane. The research in this field mainly revolves around two issues: (1) How to obtain a safe, comfort and efficient trajectory equation? (2) How to control a vehicle to accurately track the trajectory? Over the past decades, considerable efforts have been made.

The methods in trajectory planning could be roughly divided into analytical [6-8], artificial potential field [2, 9], and data-driven [4, 5]. The analytical method predetermines a trajectory form equation. The problem of obtaining corresponding equation coefficient is solved through transferring this into an optimization problem, which takes safety, comfort and efficiency into account [6-8]. Up to now, the most commonly-used mathematical equations are quintic polynomial equation, cubic polynomial equation, sine(cosine) curve equation,

This manuscript is submitted on September 20, 2021 (Corresponding author: Linbo Li). This work is supported by National Natural Science Foundation of China (grant numbers 52172331) and the National Key R&D Program of China (grant numbers 2018YFE0102800).

Yang Li, Linbo Li are with Key Laboratory of Road and Traffic Engineering of the Ministry of Education, Tongji University, 4800 Cao'an Road, Shanghai,

P.R. China (e-mail: cc960719@tongji.edu.cn, llinbo@tongji.edu.cn). Daiheng Ni is with department of Civil and Environmental Engineering, the University of Massachusetts Amherst, Massachusetts 01003, USA (ni@engin.umass.edu). He is no involvement in the research grants.

trapezoidal curve equation, Bezier curve equation, etc. Based on vehicle-to-vehicle communication, a dynamic automated algorithm was proposed in [6], which consists of trajectory planning and trajectory tracking. Bai, et al. [21] established the rectangular collision boundary to analyze the possible collision points. Yang, et al. [22] proposed a dynamic algorithm, which consists of the trajectory decision, trajectory generation, and starting-point determination module. Huang, et al. [23] incorporated the personalized driving style into the quintic polynomial function so as to meet driver's personalized lane change needs. Luo, et al. [24] absorbed the cooperative safety spacing model into a multi-vehicle cooperative automated algorithm. Chen, et al. [8] fused the cooperative safety spacing model and the prediction of leading vehicle into the algorithm. Lim, et al. [25] proposed a hybrid trajectory planning method with the strength of sampling and optimization methods.

Data-driven approaches usually refer to methods employing machine learning or deep learning algorithms, which aims to extract LC dynamics from massive trajectory data [4, 5]. Zhang, et al. [5] employed the long short-term memory model to model CF and LC behavior simultaneously. A hybrid retraining constrained training method was further established to improve the model. Xie, et al. [4] fused the deep belief network and long short-term memory together to model the process of decision and execution. Their results demonstrate that the proposed model could better reproduce LC behavior. Artificial potential field method regards various elements of driving environment, such as road edges, static obstacles, and moving obstacles as a potential energy field [2, 9]. The vehicle tries to find a trajectory with the lowest total potential field. Hang, et al. [2] combined the potential field with model predictive control model to acquire the optimal path for the vehicle. Zheng, et al. [9] established an effective trajectory planning algorithm based on the quartic Bézier curve, where the potential field functions are employed to evaluate the real-time collision risk. After obtaining the final form of the planned trajectory, the vehicle would track this curve under the controller. This may involve the determination of the vehicle model and the controller. Moreover, many scholars consider the influence of the uncertainty of the external system on the controller, and conduct the robustness analysis of the control system.

From the aforementioned literature review, it can be seen that a wealth of results has been yielded in motion planning. Nevertheless, existing research often overlooks the critically important need for the LC vehicle to complete lane-change safely while minimizing its impact on surrounding traffic flow. Numerous studies have shown that lane changes may have significant negative impact on surroundings [19, 26-28]. Such impact reflected at the micro level is the cost of each vehicle (comfort, efficiency, safety, etc.) and reflected in the macro level is the impact on traffic flow (space-mean speed, traffic density, flow). Consequently, existing algorithms might indeed be optimal for the LC vehicle, but might not be optimal for other vehicles within its area. More specifically, existing motion planning algorithms only take the comfort, safety, and efficiency of the LC vehicle in the optimization objective function, while overlooking the cost of surroundings.

This paper introduces the pareto-optimal concept into LC motion planning in order to minimize the total cost within the LC region. It is interesting and crucial research topic about how to simultaneously consider the LC vehicle and surrounding vehicles in obtaining the corresponding pareto-optimal solutions. Our algorithm not only guarantee the safety of lane change, but also minimize its impact. Admittedly, we are also aware that in some cases, the impact of lane change on traffic flow is unavoidable, so how to reduce this impact, or rather to smooth out the traffic shock wave, is of great interest and yet to be solved. The main contributions of this paper are the following: (a) we formulate a problem that simultaneously integrates the benefits of the LC vehicle and surrounding vehicles in terms of comfort, efficiency and safety. (b) we obtain the pareto-optimal solutions and front of the above optimization problem through introducing the Non-dominated sorting genetic (NSGA) -II algorithm [29]. (c) the context of this study is mixed traffic flow where Autonomous vehicles (AVs) coexists with Human-driving vehicles (HVs). This might have indispensable research implications to promote the study of LC algorithms in mixed traffic. (d) we conduct a comprehensive simulation to verify the effectiveness of our algorithm, including micro-level (trajectory), macro-level (flow, speed and density), and sensitivity analysis (scenario settings and CF model parameters).

The paper is structured as follows: Section II presents the research problem and scopes. We present the mathematical model in Section III. In Section IV, we elaborated the pareto-optimal front and NSGA-II algorithm. Section V presents the simulation experiment design and result analysis. Finally, we give the conclusion in Section VI.

## II. RESEARCH PROBLEM

Individual drivers often take actions to maximize their own interests at all times. These actions may generally be best for individuals, but not for traffic flow in their vicinity. With the continuous deepening of the application of intelligent driving environment, the advent of self-driving vehicles may provide a new means of solving this dilemma. The estimation of traffic state has been improved from the analysis based on historical data to real-time monitoring data. The vehicle to everything technology could organically connects drivers, vehicles, roads and other transportation participation elements to realize information collection, information exchange and control command execution (as shown in Fig. 2). It becomes possible for AVs to actively dissipate or reduce traffic congestion caused by LC maneuver from the source as much as possible. This may transform traffic control from conventional passive and global means to active and individualized control.

An ideal hierarchical control architecture schematic is given in Fig. 2. This architecture meets the need of real-time and large-scale calculation of vehicle-road interactive information in traffic control. The Road Side Unit (RSU) transmits the surrounding traffic environment to the AVs in real time, so that the AVs can obtain the information of the vehicle ahead in advance. The strategic level collects the information of all road sections, gives and distributes the optimization strategy at the overall level of the road network. The tactical level completes fusion and process of multi-source data in its jurisdiction, and coordinates the driving of AVs in its own jurisdiction. The operational level refers to the real-time actions and states of AVs at each moment.

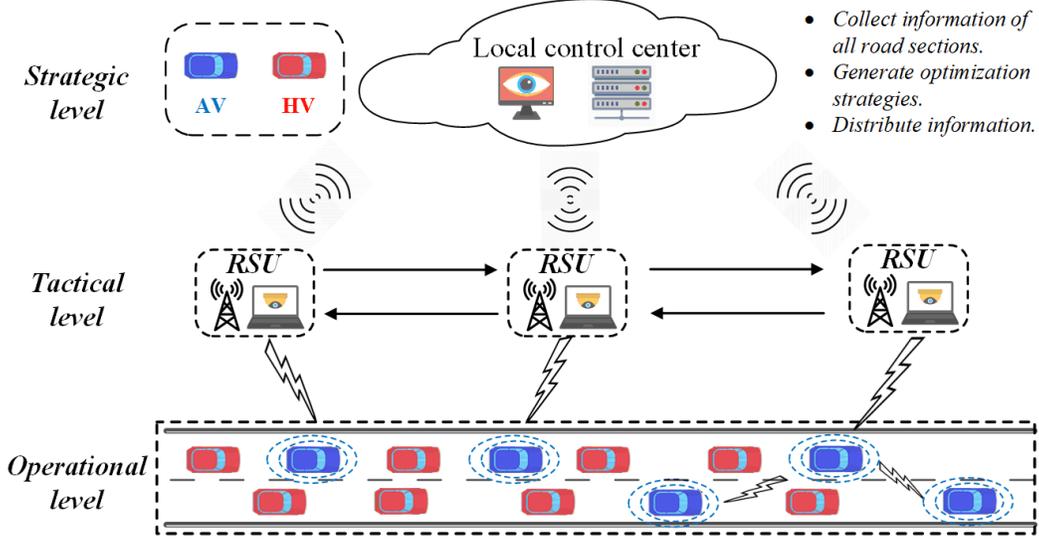

**Fig. 2 The ideal schematic diagram of the multi-level cooperative control architecture**

Traffic management agencies tend to expect lane changes to have the least impact on surroundings. As for the LC vehicles, they are more likely to maximize their own interests, and often ignore the benefits of neighboring vehicles in their strategy. Succinctly, the former tends to consider the problem from the perspective of the system as a whole, while the latter tends to start from self-interest. We assume that our algorithm could not only be distributed by the RSU to the LC vehicle, but the LC vehicle could also generate this algorithm itself (it is bound to be equipped with sensors that could obtain real-time trajectory information of surroundings). And our algorithm could also be directly translated into the algorithm commonly used in existing studies as well, by simply setting the weighting factor to zero. The major reason for showing this architecture here is to elaborate the general context in which this research is carried out. To facilitate the illustration of the subsequent study, we illustrate the constructed algorithm by standing in the perspective of RSU. Fig. 3 presents a specific LC scenario to which the algorithm applies.

Typically, the LC maneuver could be divided into two stages. One is from the decision to the execution point (stage 1). And other stage is the execution process (stage 2). In Stage 1, after $AV_{LC}$ has made the decision (at time $t_0$), it needs to search for a suitable gap on the target lane. If the gap distance is acceptable, $AV_{LC}$ will perform lane change at time $t_{start}$. Seen from the surface, $AV_{LC}$ is still in the process of CF, but in fact, $AV_{LC}$ is making preparation (finding a suitable gap). Therefore, this stage could be viewed as internal part of lane change. It is worth noting that there are also situations where $AV_{LC}$ directly executes lane change ($t_0=t_{start}$). In stage 2, the other is from the start point to the end point. $AV_{LC}$ begins to turn the steering wheel at time $t_{start}$. Then, $AV_{LC}$ gradually drives along the planned trajectory until it arrives the target lane at time $t_{end}$. This stage could be viewed as the external part. Intuitively, LC impact both exists in stage 1 and stage 2. Assume that the vehicles in the target lane maintains a stable CF time headway. Suppose the execution point is fixed, the trajectory planning algorithm is selecting one of the many curve equations that has the least cost as shown in the bottom right of Fig. 3. Each sampling point on the curve has a corresponding trajectory information, which would affect the response of other vehicles. At this stage, we need to answer whether there exists a curve that could reduce the overall cost. When the curve is fixed, the overall cost might vary under different execution point. It is even possible that the impact of the execution point is greater than the impact of the curve. Meanwhile, it might not be appropriate to research these two stages separately, since they might not be optimal at the same time.

To facilitate subsequent discussion, we need to explain the scope and the assumption: (a) only LC and CF behavior are considered, and our algorithm is more suitable for mandatory scenarios. This paper only focuses on single LC maneuver. Future research will consider multi-vehicle LC maneuver. (b) the AVs are with SAE Level 4/5 automation that could drive fully-autonomously. The AVs could obtain the real-time status of surroundings either through its own sensors, or through RSU. (c) the strategic level controller has given the range of the coordination zone to RSU in the tactical level. And the tactical level distributes an optimal execution point and trajectory information to the $AV_{LC}$. (d) $AV_{LC}$ has already determined to change lane, and the time before $t_0$ is outside the scope of this paper. Out of special reasons, $AV_{LC}$ has to perform LC. This may be due to lane-drop or low speed of preceding vehicle or the occurrence of traffic accident. This leads to a poor driving experience of $AV_{LC}$, which causes $AV_{LC}$ to perform LC.

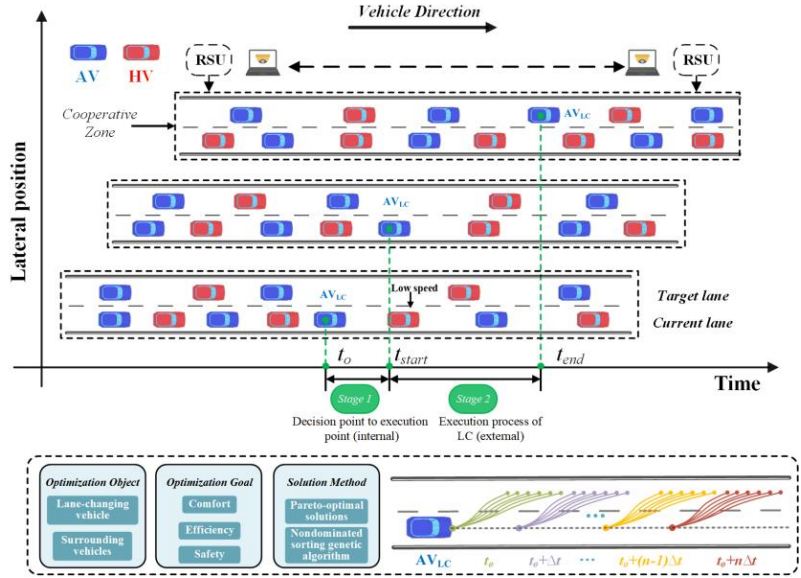

**Fig. 3 The diagram schematic of the proposed automatic LC algorithm**

## III. MATHEMATICAL MODEL

This section presents our mathematical model, consisting of the CF model, the LC model and the multiobjective optimization problem. The LCM (Longitudinal Control Model) [1, 30] is introduced to characterize the longitudinal motions of HVs. The reason why we choose this model is to the unified perspective casted on the existing microscopic traffic flow models [1]. The IDM (Intelligent Driver Model) [31] is employed to model the longitudinal motions of AVs. This model could better reproduce the CF behavior of AVs, and has been widely used as an autonomous CF model in existing research [32, 33]. As for the LC model, the time-based quintic polynomial function is employed to model the longitudinal and lateral LC trajectory, since it has high reliability and flexibility [23, 24, 34]. The multiobjective optimization function is formulated, where the benefits of $AV_{LC}$ and the surroundings are simultaneously considered. Finally, the vehicle kinematics model, vehicle error model, and the MPC controller are integrated together to track the planned trajectory.

### A. Car-following model

#### 1) LCM model

The LCM model is derived through focusing the forces for the vehicle in the longitudinal direction of Field Theory. Field Theory represents everything in the environment (highways and vehicles) as a field perceived by the subject driver whose mission is to achieve his or her goals by navigating through the overall field. The formula is given below:

$$a_{LCM_i}(t+\tau_i) = A_i[1 - \frac{v_i(t)}{v_{desire}} - e^{1-s_{ij}(t)/s_{ij}^*(t)}] \quad (1)$$

$$s_{ij}^*(t) = \frac{v_i^2(t)}{2b_i} - \frac{v_j^2(t)}{2B_j} + v_i(t)\tau_i + l_j \quad (2)$$

$$\dot{a}_{LCM_i}(t+\tau_i) = A_i[-\frac{a_i(t)}{v_{desire}} - \frac{(s_{ij}^*(t))'s_{ij}(t)}{(s_{ij}^*(t))^2} e^{1-s_{ij}(t)/s_{ij}^*(t)}] \quad (3)$$

$$(s_{ij}^*(t))' = \frac{a_i(t)}{b_i} - \frac{a_j(t)}{B_j} + a_i(t)\tau_i \quad (4)$$

Where $i$ denotes the follower vehicle, $j$ denotes the leader vehicle, $l_j$ denotes the vehicle length, $\tau_i$ denotes the reaction time, $s_{ij}(t)$ denotes the spacing between vehicle $i$ and $j$, $v_{desire}$ denotes the desired speed, $A_i$ denotes the maximum acceleration, $s_{ij}^*$ denotes the desired safe spacing of driver $i$, $b_i$ denotes the maximum deceleration that the driver can confidently apply in an emergency, $B_j$ denotes the driver's estimation of the leader vehicle's comfortable deceleration in an emergency brake.

#### 2) IDM model

The IDM (Intelligent driver model), the most popular model today, assumes that each driver has a desired headway, speed and headway, and that the model consists of acceleration in free-flow conditions and deceleration to avoid collision with the vehicle in front. The IDM model takes the following forms:

$$a_{IDM_i}(t) = A_i[1 - (\frac{v_i(t)}{v_{desire}})^\delta - (\frac{s_{ij}^*(t)}{s_{ij}(t) - l_j})^2] \quad (5)$$

$$s_{ij}^*(t) = s_{jam}^{(i)} + s_1^{(i)}\sqrt{\frac{v_i(t)}{v_{desire}}} + v_i(t)T + \frac{v_i(t)\Delta v_{ij}(t)}{2\sqrt{A_i a_i^{conf}}} \quad (6)$$

$$\dot{a}_{IDM_i}(t) = A_i[-\delta(\frac{v_i(t)}{v_{desire}})^{\delta-1} - \frac{2s_{ij}^*(t)(s_{ij}^*(t))'}{(s_{ij}(t) - l_j)^2}] \quad (7)$$

$$(s_{ij}^*(t))' = \frac{s_1^{(i)}}{2\sqrt{v_i(t)v_{desire}}} + \dot{a}_{IDM_i}(t)T - \frac{a_{IDM_i}(t)^2}{2\sqrt{A_i a_i^{conf}}} \quad (8)$$

Where $\delta$ is the acceleration coefficient, $s_{jam}^{(i)}$ is the distance between blocked vehicles, $s_i^{(i)}$ is the minimum spacing at standstill, $\Delta v_{ij}(t) = v_j(t) - v_i(t)$ denotes the relative speed. $T$ is the desired time headway, $a_i^{conf}$ is the comfortable acceleration.

### B. Lane-changing model

#### 1) Trajectory planning model

The time-based quintic polynomial function is introduced to model the longitudinal and lateral trajectory of $AV_{LC}$ [23, 24, 34]. The longitudinal and lateral trajectory with respect to time is defined as below:

$$\begin{cases} x_{LC}(t) = a_0 + a_1 t + a_2 t^2 + a_3 t^3 + a_4 t^4 + a_5 t^5 \\ y_{LC}(t) = b_0 + b_1 t + b_2 t^2 + b_3 t^3 + b_4 t^4 + b_5 t^5 \\ \theta_{LC}(t) = ar\tan\left(\dot{y}_{LC}(t) / \dot{x}_{LC}(t)\right) \end{cases} \quad (9)$$

Where $x_{LC}(t), y_{LC}(t), \theta_{LC}(t)$ denote the longitudinal position, lateral position and course angle. $a_i, i = 0,1,...5$ and $b_j, j = 0,1,...5$ are the corresponding coefficients.

Assuming time $t$, it is reasonable to assume that the speed and acceleration of $AV_{LC}$ are desired to be zero at the start and the end position in the lateral direction. Therefore, we could derive the following equations constrains.

$$\begin{cases} x_{LC}(t_{start}) = x_{start}, \dot{x}_{LC}(t_{start}) = v_{LC}^{start}, \ddot{x}_{LC}(t_{end}) = a_{LC}^{start} \\ x_{LC}(t_{end}) = x_{end}, \dot{x}_{LC}(t_{end}) = v_{LC}^{end}, \ddot{x}_{LC}(t_{end}) = a_{LC}^{end} \end{cases} \quad (10)$$

$$\begin{cases} y_{LC}(t_{start}) = 0, \dot{y}_{LC}(t_{start}) = 0, \ddot{y}_{LC}(t_{start}) = 0 \\ y_{LC}(t_{end}) = D_0, \dot{y}_{LC}(t_{end}) = 0, \ddot{y}_{LC}(t_{end}) = 0 \end{cases} \quad (11)$$

Where $D_a$ denotes the lane width, $v_{LC}^{start}$ and $v_{LC}^{end}$ denote the initial and final speed in the longitudinal direction, $a_{LC}^{start}$ and $a_{LC}^{end}$ denote the initial and final acceleration in the longitudinal direction, $x_{start}$ and $x_{end}$ denotes the initial and final longitudinal position of the $AV_{LC}$.

#### 2) Trajectory tracking model

The vehicle kinematics model is introduced. It is worth noting that the focus of this study is on the trajectory planning rather than trajectory tracking. Considering that trajectory tracking algorithm is also an indispensable part of motion planning, the model we employed in this paper is based on [35]. According to the kinematic constraints of the front and rear axles, we could derive the following equations:

$$\dot{\psi} = [\dot{x}, \dot{y}, \dot{\varphi}]^T = [\cos\varphi, \sin\varphi, \frac{1}{l}\tan\delta]^T \cdot v_r \quad (12)$$

Where $\delta$, $l$, $v_r$ denotes the front wheel steering angle, wheel base, and the speed at the center of the rear axle. In order to facilitate the design of model predictive controller, we expand this model by Taylor series at the reference trajectory point $(x_r, y_r)$, and the vehicle error model is formulated as:

$$\tilde{\psi}(k+1) = \tilde{A} \cdot \tilde{\psi}(k) + \tilde{B} \cdot \tilde{\mu}(k), (k=1,2,3\cdots) \quad (13)$$

$$\tilde{A} = \begin{bmatrix} 1 & 0 & -v_r \sin\varphi_r T \\ 0 & 1 & v_r \cos\varphi_r T \\ 0 & 0 & 1 \end{bmatrix}, \tilde{B} = \begin{bmatrix} \cos\varphi_r T & 0 \\ \sin\varphi_r T & 0 \\ \dfrac{\tan\delta_r T}{l} & \dfrac{v_r T}{l\cos^2\delta_r} \end{bmatrix} \quad (14)$$

Where $T$ denotes the sampling time, $\tilde{\psi} = \psi - \psi_r$ denotes the difference with the reference state, $\tilde{\mu} = \mu - \mu_r$ denotes the difference with the reference input, $\tilde{\psi}_r = [x_r, y_r, \varphi_r]^T$, $\mu_r = [a_r, \delta_r]^T$.

The design ideas of the controller mainly include: the current state should converge to the reference value as soon as possible, and the control input should as small as possible. Therefore, the deviation of the system state quantity and the control quantity need to be optimized. The objective function has the following form:

$$I(kq) = \sum_{i=1}^{N_p} \left\| \eta(k+i|t) - \eta_r(k+i|t) \right\|_R^2 + \sum_{i=1}^{N_c-1} \left\| \Delta U(k+i|t) \right\|_Q^2 + \rho\varepsilon^2 \quad (15)$$

Where the first item on the right side describes the rapidity of tracking control system, and the right side describes the stationary of the tracking control system. $N_p$ is the prediction horizon, $N_C$ is the control horizon, and $\rho$ is the weight coefficient. $\varepsilon$ is the relaxation factor, which could directly limit the control increment, avoid the sudden change of control quantity, and prevent the situation that there is no feasible solution in the optimization process.

### C. Multiobjective Optimization problem

#### 1) Formulation of the cost function

Suppose at time $t_0$, there are $m$ mixed vehicles behind the target lane (vehicles are marked from 1 to $m$ from near to far). The joint optimization objective involves the $AV_{LC}$ and the $m$ mixed vehicles. The optimization objective function contains comfort, efficiency and safety. The jerk variable is introduced to quantify the comfort; the difference between the current and desired speed is employed to characterize efficiency [34]; the relative speed and relative distance with preceding vehicle is used to quantify driving risk [2]. The general formulas of these three cost take the following form:

$$J_i^{comfort}(t) = \left| \dot{a}_i(t) \right| \quad (16)$$

$$J_i^{efficiency}(t) = \left| v_i(t) - v_0 \right| \quad (17)$$

$$J_i^{safety}(t) = \lambda_{safety}\left(\Delta v_{ij}(t)\right)^2 + 1 / \left[\left(s_{ij}(t)\right)^2 + v_{small}\right] \quad (18)$$

Where $J_i^{comfort}(t)$, $J_i^{efficiency}(t)$, $J_i^{safety}(t)$ denotes the comfort, efficiency and safety cost of vehicle $i$ at time $t$. $\lambda_{safety}$ equals to 1 when $\Delta v_{ij}(t) \geq 0$, and equals to 0 when $\Delta v_{ij}(t) < 0$. $v_{small}$

is a small value to avoid zero dominator. When calculating the above these three items, the following two points need to be noted. (1) if vehicle $i$ is a HV, the $\dot{a}_i(t)$ adopts the acceleration of the LCM model, and $v_0$ equals to the corresponding parameter $v_{desire}$. (2) if the vehicle $i$ is an AV, $\dot{a}_i(t)$ adopts the acceleration of the IDM model in the longitudinal direction, and adopts the acceleration of the planned trajectory in the lateral direction.

Ideally, if the speed of the target lane traffic is the same as the desired speed, the construction of Equation 17 could also be understood from the perspective of traffic shock wave. Let point B on the flow-density curve be the condition represented by the LC vehicle, and point A be the condition represented by the following vehicles in the target lane. As long as point B is on the right of point A, the shock wave will be formed eventually, and the speed of the shock wave is denoted by the slope of the chord, $U_{AB}$. If point B moves toward point A along the curve, the impact of shock wave reduces.

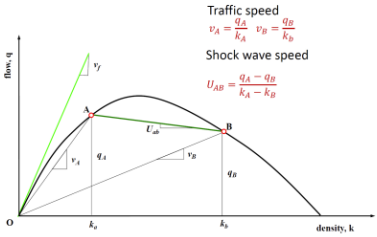

**Fig. 4 Illustration of the calculation of the traffic shock wave using the LWR model [36]**

The total cost of $AV_{LC}$ takes the following form:

$$J_{LC} = \omega_{comfort} \cdot \frac{\sum\limits_{t=t_0}^{t_{end}} J_{LC}^{comfort}(t)}{N_{comfort}} + \omega_{efficiency} \cdot \frac{\sum\limits_{t=t_0}^{t_{end}} J_{LC}^{efficiency}(t)}{N_{efficiency}}$$
$$+ \omega_{safety} \cdot \frac{\sum\limits_{t=t_0}^{t_{end}} J_{LC}^{safety}(t)}{N_{safety}} \qquad (19)$$

Where $J_{LC}$ denotes the total cost of $AV_{LC}$. During $t_0$ to $t_{start}$, the $AV_{LC}$ is controlled under the IDM model. During $t_{start}$ to $t_{end}$, the $AV_{LC}$ tracks the planned LC trajectory under the MPC controller. $N_{comfort}$, $N_{efficiency}$, $N_{safety}$ are the normalized values of the corresponding terms in the objective function (to make the units consistent). $\omega_{comfort}$, $\omega_{efficiency}$, $\omega_{safety}$ denote the weight coefficient of comfort, efficiency and safety.

The total cost of the $m$ mixed vehicles behind the target lane takes the following form:

$$J_{TF} = \omega_{comfort} \cdot \frac{\sum\limits_{i=1}^{m}\sum\limits_{t=t_0}^{t_{end}} \omega_i J_i^{comfort}(t)}{N_{comfort}} + \omega_{efficiency} \cdot \frac{\sum\limits_{i=1}^{m}\sum\limits_{t=t_0}^{t_{end}} \omega_i J_i^{efficiency}(t)}{N_{efficiency}}$$
$$+ \omega_{safety} \cdot \frac{\sum\limits_{i=1}^{m}\sum\limits_{t=t_0}^{t_{end}} \omega_i J_i^{safety}(t)}{N_{safety}}$$
$$(20)$$

Where $J_{TF}$ denotes the total cost of the $m$ mixed vehicles (TF denotes target following). $\omega_i$ denote the weight coefficient of the vehicle $i$. For vehicle with close distance and high relative speed difference, the more they are affected by the LC behavior, the larger its weight coefficient in the objective function. The $\omega_i$ takes the following form:

$$\omega_i = \sigma_i / \sum\limits_{k=1}^{m} \sigma_k \qquad (21)$$

$$\sigma_i = \left| \Delta v_{i-AV_{LC}}(t_0) \right| / \sqrt{\Delta x_{i-AV_{LC}}(t_0)} \qquad (22)$$

Where $\Delta x_{i-AV_{LC}}(t_0)$ denotes the initial distance between vehicle $i$ and $AV_{LC}$ at time $t_0$, $v_{i-AV_{LC}}(t_0)$ denotes the initial speed difference between vehicle $i$ and $AV_{LC}$ at time $t_0$.

### 2) Constrains of the problem

The $AV_{LC}$ needs to meet speed, stability, comfort, safety constrains during the execution of LC. The speed of $AV_{LC}$ should not exceed the maximum speed but should be greater than the minimum speed. The acceleration and jerk should within the reasonable range. In addition, the LC duration ($t_{end} - t_{start}$) and the longitudinal moving distance of LC ($x_{end} - x_{start}$) should also within a reasonable range. Existing research demonstrate that the LC duration roughly ranges from 1s to 16s [11, 37]. Therefore, we specify the size of the longest and shortest LCD.

$$v_{\min} \leq \sqrt{\dot{x}_{LC}(t)^2 + \dot{y}_{LC}(t)^2} \leq v_{\max} \qquad (23)$$

$$a_{\min} \leq \sqrt{\ddot{x}_{LC}(t)^2 + \ddot{y}_{LC}(t)^2} \leq a_{\max} \qquad (24)$$

$$j_{\min} \leq \sqrt{\dddot{x}_{LC}(t)^2 + \dddot{y}_{LC}(t)^2} \leq j_{\max}$$

$$0 \leq t_{start} - t_0$$
$$t_{LC,\min} \leq t_{end} - t_{start} \leq t_{LC,\max} \qquad (25)$$
$$x_{LC,\min} \leq x_{end} - x_{start} \leq x_{LC,\max}$$

Where $v_{\min}$, $v_{\max}$, $a_{\min}$, $a_{\max}$, $j_{\min}$, $j_{\max}$, $t_{LC,\min}$, $t_{LC,\max}$, $x_{LC,\min}$, $x_{LC,\max}$ represent the minimum and maximum speed limit, acceleration, jerk, LC duration, longitudinal moving distance.

The $AV_{LC}$ must not collide with its surrounding vehicles at any time. The definition of the collision boundary area is shown below. $l_a, l_b, C_a, C_b$ are defined as vehicle length, vehicle width, ellipse long radius and ellipse short radius respectively.

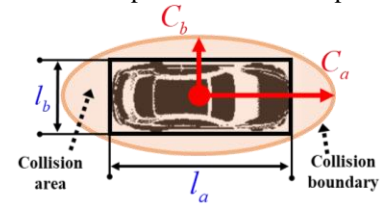

**Fig. 5 The boundary of the collision area of the LC vehicle**

Taking the starting point as the original coordinates, suppose at time $t$, let $P_{AV_{LC}}(t)=\left(x_{AV_{LC}}(t),y_{AV_{LC}}(t)\right)$ denotes the center position of the vehicle $AV_{LC}$. The real-time boundary of collision area of $AV_{LC}$ is defined as $G_{AV_{LC}}(x,y)$.

$$\begin{cases} M^2/C_a{}^2+N^2/C_b{}^2=1 \\ M=\left(x-x_{AV_{LC}}(t)\right)*\cos\theta_{AV_{LC}}-\left(y-y_{AV_{LC}}(t)\right)*\sin\theta_{AV_{LC}} \\ N=\left(x-x_{AV_{LC}}(t)\right)*\sin\theta_{AV_{LC}}+\left(y-y_{AV_{LC}}(t)\right)*\cos\theta_{AV_{LC}} \end{cases} \quad (26)$$

It is worth noting that the four corners of the smallest circumscribed rectangle of the vehicle outline should fall on the ellipse or within the ellipse. The real-time minimum distance between two collision boundaries could be obtained through the Lagrangian solution algorithm [34].

## IV. SOLUTION METHOD

### A. Pareto optimality

To reduce the impact of LC on its surroundings, we convert this problem into a pareto optimality problem. Pareto optimality is a situation where no individual or preference criterion can be better off without making at least one individual or preference criterion worse off or without any loss thereof [38]. Our problem of minimizing $J_{TF}$ and $J_{LC}$ can be converted into multiobjective optimization problem, where we can only find a set of acceptable solutions [39]. Since the multiobjective model does not has a universal optimal solution, it is more important to find good compromises, or trade-offs, rather than a single solution.

Let $\Gamma$ be the set of all feasible solutions. $x_A$, $x_B$, $x_C$, $x_D$, $x_E$ denote five different feasible solutions within the hypothetical feasibility region as shown in Fig. 6. If the value of the objective function corresponding to solution A is better than the value of the objective function corresponding to solution B, it could be called solution A strongly Pareto dominance solution B (for example, solution $x_E$ is better than solution $x_D$ and solution $x_E$). If an objective function value corresponding to solution A is better than an objective function value corresponding to solution B, but another objective function value corresponding to solution A is worse than an objective function value corresponding to solution B, then solution A is indistinguishable from solution B, also known as solution A can Pareto dominate solution B (for example the relation between $x_D$ and $x_E$). For solution A, if no other solution can be better than solution A in the variable space $\Gamma$ (both objective function values are better than the function value corresponding to A), then solution A is the pareto-optimal solution (for example the solution $x_A$ and $x_B$). The solid dots in Fig. 6 are all pareto-optimal solutions. All pareto-optimal solutions constitute the pareto-optimal solution set, and these solutions are mapped by the objective function to form the pareto-optimal front of the problem.

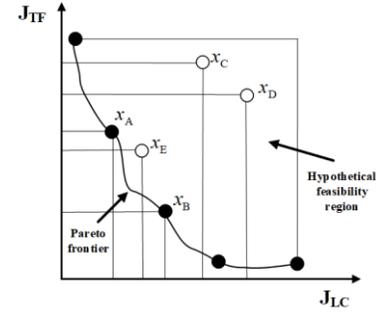

**Fig. 6 The illustration of Pareto-optimal solutions**

### B. NSGA-II algorithm

Over the past decades, multiobjective GA (Genetic Algorithm) [29, 40-43], as one kind of evolutionary algorithm, has been extensively employed to solve the multiobjective optimization problem. Various kinds of multiobjective GA has been proposed, such as vector evaluated GA [40], non-dominated sorting GA [41], random weight GA [42], adaptive weight GA[43], non-dominated sorting GA II [29], etc. Among these algorithms, the most famous is the NSGA-II algorithm, which has been widely applied in the field of transportation[44-47]. Compared with the first version of NSGA, NSGA-II algorithm has lower computational complexity. Crowding degree and comparison operators are introduced to maintain the diversity of populations. The elite strategy is adopted to expand the sampling space and improve the population level rapidly [29]. For more details of NSGA-II, please refer to [29].

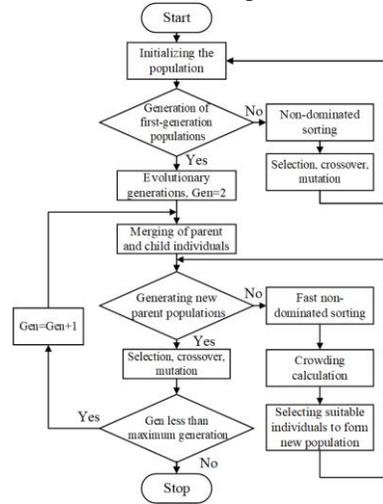

**Fig. 7 Basic process of the NSGA-II algorithm**

The basic process of the NSGA-II algorithm is presented in Fig. 7. Three main steps are summarized below. Step 1: The initial population of size is randomly generated, and the first generation of offspring population is obtained by the three basic operations of selection, crossover and mutation of genetic algorithm after non-dominated sorting. Step 2: From the second generation, the parent population is merged with the offspring population to perform fast non-dominated sorting, and the crowding degree is calculated for the individuals in each non-dominated layer. Based on the non-dominance relationship and the crowding degree of the individuals, the appropriate individuals are selected to form the new parent population. Step 3: A new population of children is generated by the basic operations of the genetic algorithm, and so on until the end of the program is satisfied. For more details about this algorithm, please refer to [29].

### C. Testing environment

All the simulation is programmed in Python, and run on an i9-9700CK 3.6GHz processor, with 16GM RAM and RTX 2070. The NSGA-II algorithm in pymoo package is employed to solve the multiobjective problem. This package is developed under the supervision of Kalyanmoy Deb [29]. In order to improve the speed of the algorithm solution, we define the magnitude of change for each variable each time. We set the minimum change unit of each variable to 0.1, since too many decimal places only increase the complexity of the solution, and do not make a fundamental difference to the overall LC maneuver (for example, the LC duration takes the values of 5.11 and 5.12, or the final speed takes the values of 24.51 and 24.52). This could be realized in the pymoo package.

## V. SIMULATION EXPERIMENT DESIGN AND RESULTS ANALYSIS

### A. Scenario and parameters settings

The simulation step $\Delta t$ is set as 0.1s. The parameters of the CF and LC models need to be set in advance for the numerical simulation. Thus, this study borrows some parameters from the existing literature, whose values are often used in existing studies. The parameters of the LCM and IDM models are from [48, 49]. The normalized values of the corresponding cost terms are set as: $N_{comfort} = 8m/s^3$, $N_{efficiency} = 25m/s$, $N_{safety} = 0.5s^{-1}$. Other paraemters are set as: $l_a = 5m$, $l_b = 2m$, $C_a = 2.5m$, $C_b = 1m$, $v_{max} = 30m/s$, $v_{min} = 5m/s$, $a_{max} = 8m/s^2$, $a_{min} = -8m/s^2$, $j_{max} = 8m/s^3$, $j_{min} = -8m/s^3$, $D_o = 3.5m$ [34]. The simulation duration is set to 500s, with a total of 20 mixed vehicles on the target lane. We consider the scenario of an accident ahead (it is equivalent to a lane drop in front of the vehicle). We assume that the current lane ends, and the vehicle has to make a lane change. After a period of warm-up simulation time, the vehicles on the target lane all maintain a steady CF distance and speed.

Our simulation scenarios are randomly generated to ensure that the algorithm applies to all possible traffic conditions. The following scenarios is one of them with penetration rate as 50%, the AV and HV alternatively appear in the queue. Suppose at moment 300s, the longitudinal position of $AV_{LC}$ is between the 10th and the 11th vehicle. At this moment, two vehicles 100m ahead of $AV_{LC}$ suddenly collide and both vehicles stop abruptly. The initial speed of $AV_{LC}$ is 20m/s, the distance to the 11th vehicle to 20m (sensitivity analysis for these parameters will be conducted in the next subsection). The $AV_{LC}$ has to perform lane-change between the 10th and the 11th vehicle.

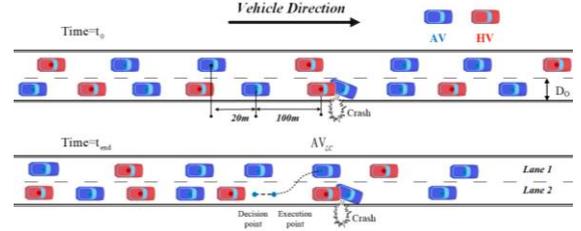

**Fig. 8 The illustration of the testing scenario**

### B. Performance of our proposed algorithm

Fig. 9 presents the pareto-optimal front. With the origin as the center, a circle is drawn with the closest distance of the frontier from the origin (the radius is 36.81), and the intersection of the circle and the frontier is taken as the final optimal solution. Since the construction of the cost function for $AV_{LC}$ and the mixed vehicles in Section 3 is the same, we could take such an intersection point[50] as our selected solution. Under this solution, the cost of $AV_{LC}$ is about 14.11, the cost of mixed vehicles is 34.00, and the total cost is about 48.11. When the cost of $AV_{LC}$ is rather small, the cost of mixed vehicles is extremely large. For example, the leftmost scatter point, when the cost of $AV_{LC}$ is less than 3, the cost of mixed vehicles is all greater than 50, and the total cost is all greater than the cost corresponding to the intersection we take. The total cost is reduced by 12.22%.

Subplot (b), (c), (d) in Fig. 9 presents the longitudinal trajectories (position, speed and headway) of all vehicles. The red curve represents the $AV_{LC}$, the blue curve represents the AVs, and the black curve represents the HVs. After a period of simulation warm-up time, the 20 mixed vehicles maintain a constant CF time headway. The time headway of AVs is lower than that of the HVs (about 2.4s for AV and 3.2s for HV). When the $AV_{LC}$ is inserted into the 10th and 11th vehicle, its LC behavior significantly affected the mixed vehicles behind (all of them made corresponding deceleration).

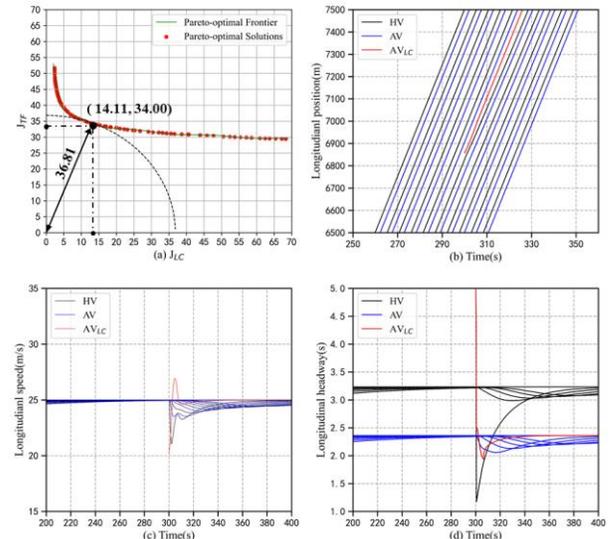

**Fig. 9 Pareto-optimal front, longitudinal trajectories of all vehicles**

Fig. 10 presents a more detailed microscopic view of the movement and cost of each vehicle. We demonstrate the longitudinal and lateral trajectory of $AV_{LC}$, the longitudinal trajectories of three vehicles behind, and the cost of each vehicle at each moment. At the moment of 300s, the $AV_{LC}$ has decided to change lane, but it does not choose to steer the wheel at this moment, but until the moment of 300.6s. And the $AV_{LC}$ finally reached the centerline of the target lane at 306.6s. Within 300s to 300.6s, $AV_{LC}$ still travels according to the IDM model, and three vehicles behind the target lane also travel according to the steady CF speed. The speed of $AV_{LC}$ is 20m/s at the moment of 300s. The speed of $AV_{LC}$ is 20.5m/s, and the acceleration is about $0.81 \text{m/s}^2$ when time is 300.6s.

Within 300s to 300.6s, $AV_{LC}$ follows the trajectory equation given by our algorithm, and the 11th vehicle, 12th vehicle, 13th vehicle gradually decelerate. At the moment of 306.6s, the longitudinal speed of $AV_{LC}$ arrives at 26m/s, and the longitudinal acceleration reaches at 0. At the same time, it is worth noting that the speed and acceleration of the $AV_{LC}$ are all continuous everywhere during the period from stage 1 (from 300s~300.6s) to stage 2 (from 300.6s to 306.6s). Subplot (c) and Subplot (d) also present the lateral trajectory information of the $AV_{LC}$. The lateral trajectory of $AV_{LC}$ smoothly and gradually changes from 0m to 3.5m, and the lateral speed slowly increases from 0m to 1.1m/s and gradually decreases to 0. The four subplots at the bottom of Fig. 10 presents the total cost, comfort cost, efficiency cost and safety cost for these four vehicles for each moment. At moment 300.6s, the instantaneous comfort cost of the 11th vehicle exceeds 20, while its safety cost is 18.12, resulting in a total cost of nearly 25 at that moment, causing it to be the one vehicle most affected.

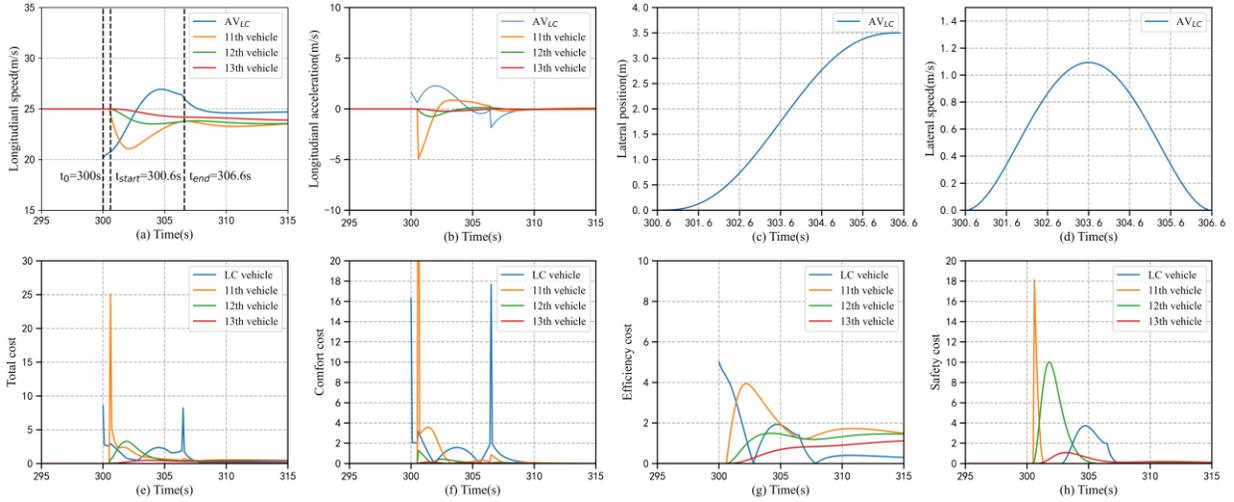

**Fig. 10 Lateral and longitudinal trajectories of the $AV_{LC}$, and the trends of cost function for each vehicle**

### C. Comparison with the existing algorithm

When $AV_{LC}$ takes the leftmost solutions, the cost of $AV_{LC}$ could indeed have a lower cost, but the corresponding cost of the mixed vehicles is larger, leading to a not-optimal overall cost. In this subsection, we introduce a macro-level analysis approach in this subsection to compare and analyze the impact of these two situations on traffic flow. To facilitate the presentation of our macro-level analysis, we increase the number of vehicles in the target lane to 200, with $AV_{LC}$ changing lanes between the 100th and 101th vehicles. It is worth noting that we did not change the hourly traffic flow. We merely increase the number of vehicles in the simulation to facilitate the observation of the traffic shock waves induced by the LC behavior at the macro level.

We increase the simulation step size to 6000s (enables all vehicles to reach a stable CF time headway), and the $AV_{LC}$ decides to change lane at 4000s, keeping all the remaining parameters the same as in the previous subsection. The TTT (Total travel time) metric is also introduced to assess the traffic operation and efficiency of transportation system. Through comparing the TTT difference for all vehicles to pass through the downstream road section under the two situations where the AV changes lane and the AV does not change lane, we could evaluate the impact of LC maneuver. The cross section is selected 200 meters downstream from the starting point of lane-change, and we evaluate the total difference of the rear 20 vehicles.

Since the speed of $AV_{LC}$ during LC is not fixed, it is not appropriate for us to estimate the traffic shock wave using the steady-state macroscopic model. Nevertheless, we could employ the basic method to explore the flow, space-mean speed, and density [51, 52] through defining the rectangular area within the LC area. This method takes the following three forms:

$$q_{flow} = d(A)/|A| \qquad (27)$$

$$V_{space-mean} = d(A)/t(A) \qquad (28)$$

$$k_{density} = t(A)/|A| \qquad (29)$$

Where $|A|$ denotes the area of rectangle, $d(A)$ and $t(A)$ denotes the total distance and time travelled of all vehicles within this area.

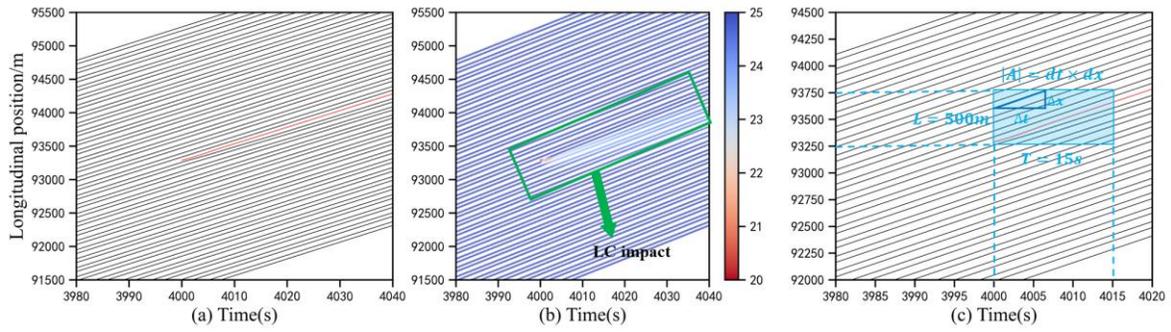

**Fig. 11 Longitudinal speed time-space heatmap of the vehicles within the LC area**

**Tab. 1 Time and distance travelled of each vehicle within the LC area, and the corresponding macroscopic traffic flow index**

|  | The existing algorithm | | Our proposed algorithm | |
|---|---|---|---|---|
|  | Distance travelled(m) | Time travelled(s) | Distance travelled(m) | Time travelled(s) |
| 94th vehicle | 5.00 | 0.30 | 5.00 | 0.30 |
| 95th vehicle | 85.00 | 3.50 | 85.00 | 3.50 |
| 96th vehicle | 140.00 | 5.70 | 140.00 | 5.70 |
| 97th vehicle | 220.00 | 8.90 | 220.00 | 8.90 |
| 98th vehicle | 275.00 | 11.10 | 275.00 | 11.10 |
| 99th vehicle | 355.00 | 14.30 | 355.00 | 14.30 |
| 100th vehicle | 375.00 | 15.10 | 375.00 | 15.10 |
| $AV_{LC}$ | 368.67 | 15.10 | 370.77 | 15.10 |
| 101th vehicle | 343.41 | 15.10 | 346.37 | 15.10 |
| 102th vehicle | 305.26 | 13.10 | 309.79 | 13.20 |
| 103th vehicle | 236.84 | 10.00 | 238.31 | 10.00 |
| 104th vehicle | 184.95 | 7.70 | 185.74 | 7.70 |
| 105th vehicle | 110.57 | 4.60 | 110.89 | 4.60 |
| 106th vehicle | 56.98 | 2.40 | 57.07 | 2.40 |
| Total | 3061.68 | 126.90 | 3073.93 | 127.00 |
| Flow (veh/h) | 1469.61 | | 1475.49 | |
| Speed (m/s) | 24.13 | | 24.20 | |

Fig. 11 presents the speed time-space heatmap of vehicles within the LC area. The abscissa represents the simulation time, the ordinate represents the longitudinal position, and each curve represents the trajectory of the vehicle. The red line in subplot (a) represents the LC vehicle. The closer the color is to blue, the higher the speed of vehicles. The closer the color is to red, the lower the speed of vehicles. The blue area represents the speed around 22.5~25m/s, the red area represents the speed around 20~22.5 m/s. The color within the LC area gradually shifts from dark blue to light blue, where the color of the LC vehicle and its immediate rear line both show red. Through the speed time-heatmap in subplot (b), we could obviously find that, with the insertion of the LC vehicle, the speed of the vehicles behind the target lane all made a corresponding deceleration.

To further analyze the LC impact, we extract the trajectories within the blue box as shown in subplot (c). Tab. 1 presents the comparison of the time and distance travelled of each vehicle within the LC area under our proposed algorithm and the existing algorithm. According to the method in [51], the flow, space-mean speed, and density could be estimated. $|A|$ equals to $L \times T = 500 \times 15 = 7500 m \cdot s$. In total, there are 14 vehicles in the region, from the 94th to 106th vehicle. It can be seen that the distance and time travelled of the previous vehicles including the 100th itself are the same under the two different algorithms. And with the insertion of $AV_{LC}$, the trajectories of the vehicles after the 100th have a significant difference. Under our algorithm, the 101th, 102th and 103th vehicles travel almost 3m, 4m and 2m more in this area. This results in an improvement in space-mean speed and traffic flow, with an overall speed increase of 0.7m/s and a flow increase of 5.88veh/h. The total difference under our algorithm and the benchmark algorithm is 25.58s and 28.71s, respectively. It has been demonstrated that our approach does, in comparison to the approach on the front's leftmost side, improve the functioning of the regional traffic flow.

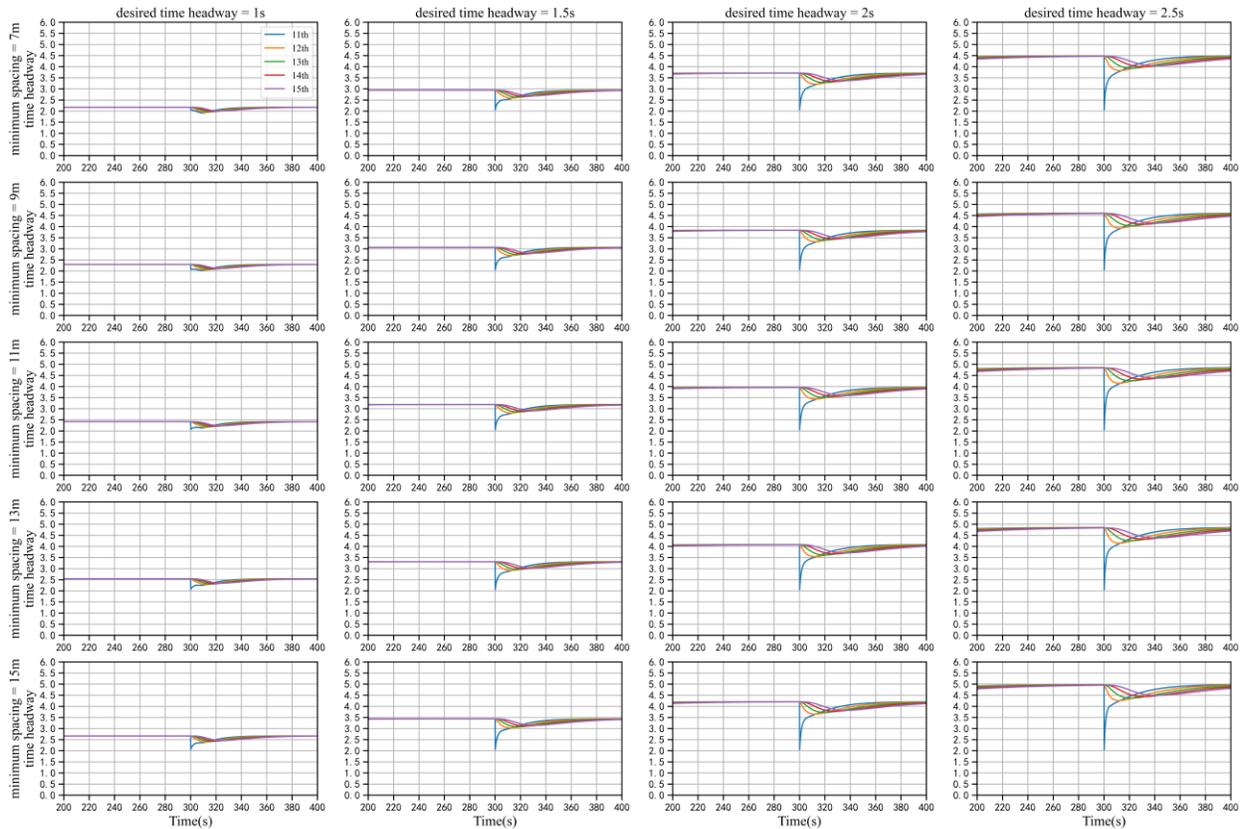

**Fig. 12 Time headway of the AVs under different minimum spacing and desired time headway regimes**

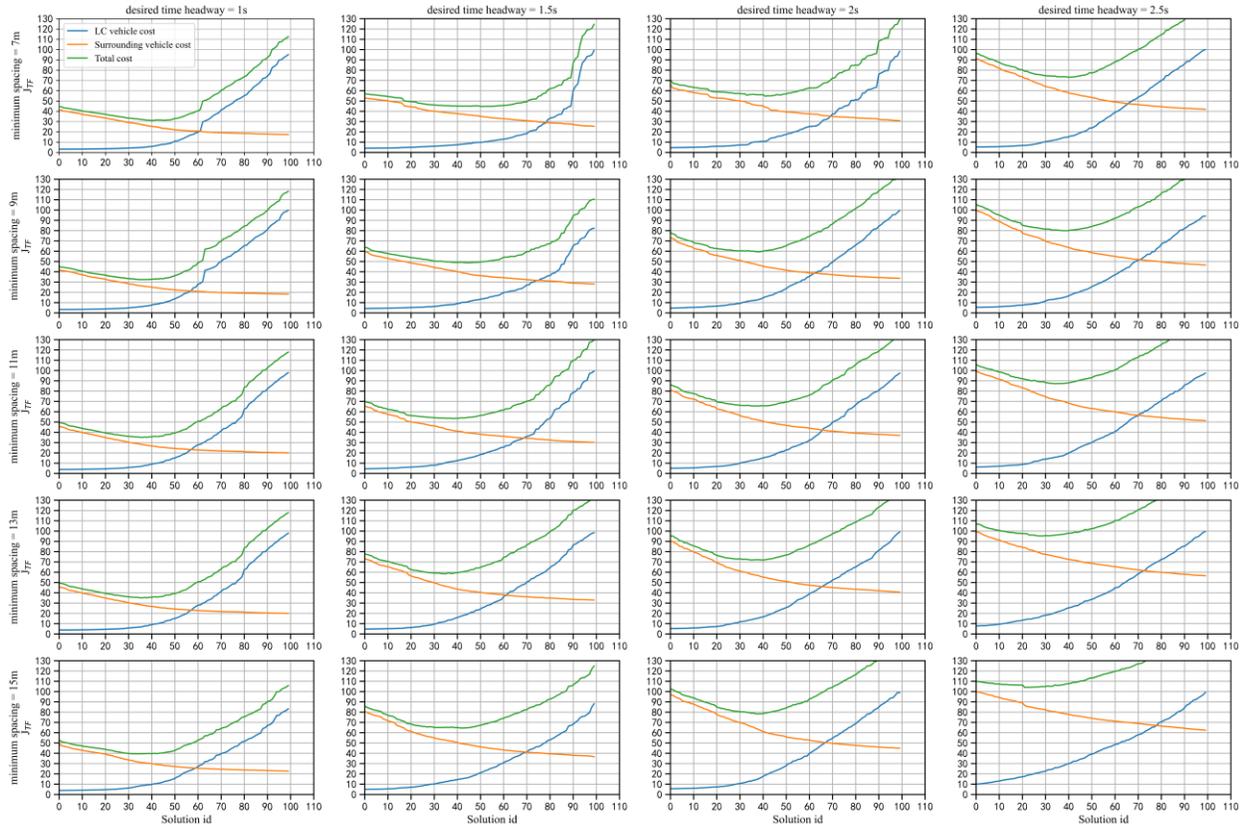

**Fig. 13 LC vehicle, surrounding vehicle and total cost under different minimum spacing and desired time headway regimes**

### D. Sensitivity Analysis of the parameters

In this section, we carry out a thorough sensitivity analysis. The analysis of the initial speed and the separation from the eleventh vehicle are shown in Fig. 14. The pareto front drops down as the initial speed increases from 18 m/s to 25 m/s. As the initial distance increases from 40m to 100m, the front gradually moves downward and the overall cost gradually decreases. Regarding the change in the final solution, we can visually see that the total cost of the vehicle behind the target lane increases significantly. When the speed is 25 m, the total cost is within 30 (tangenting the front with a circle). While when the speed is reduced to 18 m, the total cost has exceeded 80. The most important reason for this change in the final solution is the vehicle behind the target lane. This is in line with our common sense, as the insertion speed decreases or the distance decreases, the greater the impact will be.

Further, we vary the driving strategies of AVs, and analyze the effect of its change on the front. AVs may choose a cautious strategy or an aggressive one. Since there are not many vehicles behind the target lane, it's tough to anticipate the front changing as the penetration steadily grows. Future work will explore how to set different permutations at different permeabilities. Therefore, we vary the time headway of AVs and compare the performance of existing algorithms with our algorithm. Fig. 12, Fig. 13 and Tab. 2 presents the result of the sensitivity analysis for different driving strategies of AVs. We set the minimum spacing at standstill that ranges from 7m to 15m, and desired time headway that ranges from 1s to 2.5s. As the minimum spacing at standstill or desired time headway increased as shown in Fig. 12, the time headway exhibited a gradual upward trend. Correspondingly, the total cost curve in Fig. 13 is also gradually shifted upward. In Fig. 13 and Tab. 2, while the LC vehicle tries to maximize its own benefit, the overall benefit within the LC area cannot be reduced. And there is indeed such a local optimal solution, which makes the overall benefit in the region reach the maximum value. From Tab. 2, we could clearly observe that the total cost within the observed region is indeed reduced. When minimum spacing is 11m and the desired time headway is 1s, the total cost is reduced by 48.66%. At the same time, the total cost shows an increasing trend as the time headway keeps increasing.

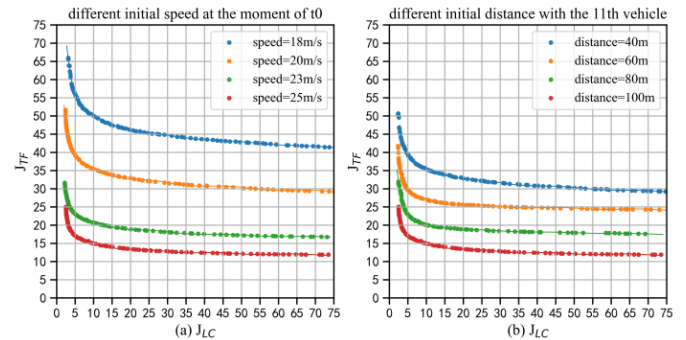

**Fig. 14 Sensitivity analysis of the front under different simulation parameters**

**Tab. 2 Comparison of the existing algorithm and our proposed algorithm**

| Desired time headway | Minimum spacing | Time headway | The existing algorithm cost | | | Our proposed algorithm cost | | | Improvement |
|---|---|---|---|---|---|---|---|---|---|
| | | | LC vehicle | Surrounding vehicle | Total | LC vehicle | Surrounding vehicle | Total | |
| 1s | 7m | 2.17s | 3.19 | 41.36 | 44.55 | 5.82 | 25.31 | 31.13 | 43.11% |
| | 9m | 2.29s | 3.34 | 41.80 | 45.14 | 6.21 | 26.21 | 32.42 | 39.24% |
| | 11m | 2.42s | 3.88 | 45.67 | 49.55 | 7.12 | 26.21 | 33.33 | 48.66% |
| | 13m | 2.54s | 3.88 | 45.66 | 49.54 | 7.12 | 28.03 | 35.15 | 40.94% |
| | 15m | 2.66s | 3.91 | 48.99 | 52.90 | 7.54 | 32.01 | 39.55 | 33.75% |
| 1.5s | 7m | 2.94s | 4.12 | 52.82 | 56.94 | 9.82 | 34.90 | 44.72 | 27.33% |
| | 9m | 3.06s | 4.35 | 59.29 | 63.64 | 11.12 | 37.63 | 48.75 | 30.54% |
| | 11m | 3.19s | 4.56 | 65.65 | 70.21 | 11.53 | 41.68 | 53.21 | 31.95% |
| | 13m | 3.31s | 4.79 | 72.96 | 77.75 | 12.28 | 46.17 | 58.45 | 33.02% |
| | 15m | 3.43s | 5.03 | 80.81 | 85.84 | 15.21 | 49.26 | 64.47 | 33.15% |
| 2s | 7m | 3.25s | 4.57 | 66.28 | 70.85 | 10.80 | 43.81 | 54.61 | 29.74% |
| | 9m | 3.37s | 4.80 | 73.29 | 78.09 | 12.90 | 46.52 | 59.42 | 31.42% |
| | 11m | 3.50s | 5.03 | 81.20 | 86.23 | 13.73 | 51.80 | 65.53 | 31.59% |
| | 13m | 3.62s | 5.30 | 89.95 | 95.25 | 16.55 | 55.32 | 71.87 | 32.53% |
| | 15m | 3.74s | 5.57 | 97.10 | 102.67 | 14.34 | 63.97 | 78.31 | 31.11% |
| 2.5s | 7m | 3.71s | 5.28 | 90.86 | 96.14 | 15.18 | 57.84 | 73.02 | 31.66% |
| | 9m | 3.83s | 5.54 | 99.84 | 105.38 | 15.18 | 64.72 | 79.90 | 31.89% |
| | 11m | 3.95s | 6.17 | 99.53 | 105.70 | 15.72 | 71.55 | 87.27 | 21.12% |
| | 13m | 4.01s | 7.76 | 99.30 | 107.06 | 17.10 | 78.18 | 95.28 | 12.36% |
| | 15m | 4.21s | 10.10 | 99.79 | 109.89 | 17.59 | 81.46 | 99.05 | 10.94% |

## VI. Conclusion

This paper introduces for the first time the concept of pareto optimal planning strategies with respect to lane change in the context of AV vehicles in mixed driving environments. We decompose the lane change into two stages. In order to simulate the scene of mixed traffic, we introduce the LCM and IDM model to characterize the longitudinal movements of HVs and AVs. We construct the cost functions for the LC vehicle and the surrounding mixed vehicles. Since the multiobjective problem does not has a universal optimal solution, it is more important to find good compromises, or trade-offs, rather than a single solution. Therefore, we introduce the NSGA-II algorithm to solve our problem. After obtaining the pareto-optimal front, we take the origin as the center and draw a circle to take the intersection point on the tangent line with the front as the final optimal solution.

Through comprehensive numerical simulation, we have verified the validity of the algorithm. Our numerical simulation is divided into three parts: the analysis of the vehicle cost function from a micro perspective; the analysis of the traffic flow operation state within the LC area from a macro perspective; the sensitivity analysis of the simulation scene parameters and the driving strategy of the AVs. With our algorithm, the impact of lane change on surrounding flow is diminished as much as possible. This is reflected in the reduction of the total cost of surrounding vehicles and the increase of traffic flow within the investigated region. It is possible to draw a line along the surface that is infinitely close to the front, but at the same time parallel to the X-axis. The intersection of this line with the Y-axis could be identified as "baseline impact". It is understood that whatever LC strategy the vehicle adopts at that point, it would incur cost of at least this size. The magnitude of this cost may be influenced to a large extent by the initial vehicle speed, as well as the distance to the first rear vehicle on the target lane. Our results show a 12.22% drop in the cost of all vehicles within the LC area, and the traffic flow within this area has increased by 6 veh/h in the first simulation scenario. Further, we comprehensively analyze the impact of the driving strategies of AVs on pareto front. Our findings show that the driving strategy of AVs does affect the position of the pareto front. As the time headway continues to increase, the overall cost within the LC region increases. Tab. 2 gives further information on each cost. We can also intuitively observe that a locally optimal solution does exist within the LC region. The total cost in the region can indeed be further reduced under our algorithm. With our algorithm, the total cost in the region can be reduced by up to 48.66%.

Undoubtedly, many aspects of this paper need further research. First, this paper makes simplifications for the construction of the cost function (safety, comfort and efficiency). For example, comfort and acceleration may have a non-linear relationship, and the driver does not always feel uncomfortable when accelerating. Future studies could further introduce probabilistic approaches to refine the modeling of the cost function. Meanwhile, multiple vehicles LC scenario will also be refined in a subsequent study. Second, subsequent research could consider how to test our algorithm using field data, minimize the errors between the simulated data and field data, and determine the parameter values in our algorithm (normalized values, maximum and minimum threshold values). For example, in some traffic simulation software, we could give a set of parameter values in advance, and then users can set them by themselves. Third, the trade-off between efficiency, safety, and comfort involves the study of another pareto-optimal front. Due to page limit, we plan to carry out this work in a follow-up study.

### Acknowledgment

The authors would like to thank the Associate Editor and anonymous reviewers for their insightful and constructive comments that helped them improve this work greatly. Daiheng Ni is not involved in the research grants.

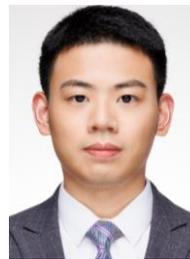

**Yang Li** was born in Shanghai, China. He received the B.S. degree in traffic engineering from Dalian Maritime University, Dalian, Liaoning, China, in 2018. He is currently pursuing the Ph.D. degree with Tongji University. His main interests include traffic flow theory and simulation, connected vehicle technology, and traffic data analysis.

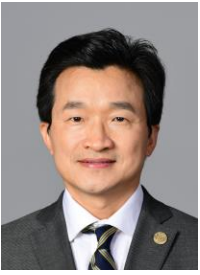

**Linbo Li** is Associate Professor of Transportation Engineering at Tongji University, CN. He received the Ph.D. degree in Transportation Engineering from Tongji University in 2007. His research interests include traffic flow theory, traffic planning and management, and smart parking. He is the author of two books and more than 30 refereed journal articles and was granted 6 national invention patents.

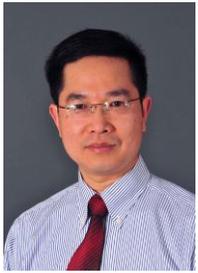

**Daiheng Ni** is Professor of Civil and Environmental Engineering at University of Massachusetts Amherst, USA. He received the Ph.D. degree in Civil Engineering from Georgia Institute of Technology, GA, in 2004. His research interests include traffic flow theory and simulation, connected vehicle technology, and Intelligent Transportation Systems (ITS). He is the author of two books and more than 50 refereed journal articles. He is an Associate Editor of Journal of Intelligent Transportation Systems, Taylor & Francis.